\documentclass[DIV=11,a4paper]{artclcomp}

\usepackage{amsmath}
\usepackage{amssymb}
\usepackage{times}
\usepackage{calc}
\usepackage{enumitem}
\usepackage{eurosym}
\usepackage{textcomp}
\usepackage{mathrsfs}
\usepackage{graphicx}
\usepackage{enumitem}
\usepackage{stmaryrd}
\usepackage{epsfig}
\usepackage[usenames,dvipsnames]{xcolor}

\usepackage{theoremref} %bessere Verweise bei Thm

\newcommand{\R}{\mathbb{R}}

\usepackage[hyperref,amsmath,thmmarks]{ntheorem}

\theoremseparator{.}
\theoremsymbol{\ensuremath{\blacklozenge}}
\theoremheaderfont{\itshape}

\numberwithin{equation}{section}

\usepackage{lipsum}
\usepackage{graphicx}
\usepackage{amsfonts}
\usepackage{accents}
\usepackage{xspace}
\usepackage{dsfont}
\usepackage{bbm} %u.a. Indikatorfunktion
\usepackage{subcaption}
\usepackage{theoremref}
\usepackage[mathscr]{euscript}
\usepackage{float}
\usepackage[format=plain,labelsep=period,labelfont=bf]{caption}

\usepackage[round,sort,comma]{natbib}

\allowdisplaybreaks[1]

\def\br{\ensuremath{\boldsymbol{r}}}

\def\bx{\ensuremath{\boldsymbol{x}}}

\def\ba{\ensuremath{\boldsymbol{a}}}

\DeclareMathAccent{\what}{\mathord}{largesymbols}{"62}

\DeclareFontFamily{U}{mathx}{\hyphenchar\font45}
\DeclareFontShape{U}{mathx}{m}{n}{
      <5> <6> <7> <8> <9> <10>
      <10.95> <12> <14.4> <17.28> <20.74> <24.88>
      mathx10
      }{}
\DeclareSymbolFont{mathx}{U}{mathx}{m}{n}
\DeclareFontSubstitution{U}{mathx}{m}{n}
\DeclareMathAccent{\widecheck}{0}{mathx}{"71}

\begin{document}

\title{Sparse Explanations of Neural Networks Using Pruned Layer-Wise Relevance Propagation}

\author[a,1]{Paulo Yanez Sarmiento}
\author[a,2]{Simon Witzke}
\author[b,3]{Nadja Klein}
\author[a,4,s]{Bernhard Y. Renard}

\address[a]{Hasso Plattner Institute, Digital Engineering Faculty, University of Potsdam, Germany}
\address[b]{Scientific Computing Center, Karlsruhe Institute of Technology, Germany}
\eMail[1]{Paulo.Yanez@hpi.de}
\eMail[2]{Simon.Witzke@hpi.de}
\eMail[3]{Nadja.Klein@kit.edu}
\eMail[4]{Bernhard.Renard@hpi.de}

\myThanks[s]{Corresponding author}

\abstract{
Explainability is a key component in many applications involving deep neural networks (DNNs). However, current explanation methods for DNNs commonly leave it to the human observer to distinguish relevant explanations from spurious noise. This is not feasible anymore when going from easily human-accessible data such as images to more complex data such as genome sequences. To facilitate the accessibility of DNN outputs from such complex data and to increase explainability, we present a modification of the widely used explanation method layer-wise relevance propagation. Our approach enforces sparsity directly by pruning the relevance propagation for the different layers. Thereby, we achieve sparser relevance attributions for the input features as well as for the intermediate layers. As the relevance propagation is input-specific, we aim to prune the relevance propagation rather than the underlying model architecture. This allows to prune different neurons for different inputs and hence, might be more appropriate to the local nature of explanation methods. To demonstrate the efficacy of our method, we evaluate it on two types of data: images and genome sequences. We show that our modification indeed leads to noise reduction and concentrates relevance on the most important features compared to the baseline.
}

\keyWords{deep learning; explainable AI (XAI); genomics; pruning; sparsity}

% \ArXiV{}

\acknowledgement{We gratefully acknowledge funding by grants KL 3037/7-1 (to NK) and RE 3474/8-1 (to BYR), project P5 in the Research Unit KI-FOR 5363 (grant 459422098) of the German Research Foundation (DFG). First published in \textit{Machine Learning and Knowledge Discovery in Databases. Research Track. ECML PKDD 2024. Lecture Notes in Computer Science}, Vol. 14944, pp 336--351, 2024 by Springer Nature.}

\date{} \maketitle \frenchspacing 

%% ====================================================================================================================================================================================== %%
%% ====================================================================================================================================================================================== %%

\section{Introduction}\label{sec_introduction}
As the usage of deep neural networks (DNNs) has grown tremendously in a variety of fields, so has the interest in the explainability and interpretability of such models. Especially for sensitive areas such as medical imaging \citep{litjens2017survey} or genomics \citep{eraslan2019deep, bartoszewicz2021interpretable}, an understanding of how the model came to certain predictions is essential to build and ensure trustworthiness. Therefore, so-called post-hoc attribution or explanation methods have been introduced \citep{samek2021explaining} to decode the black-box behavior common for DNN architectures. Thereby, we assume there is an already trained model and consider a separate method to explain it. Instead of global explanations of the model, our focus is on local methods. Their general idea is to obtain an input-specific explanation of the decisive behavior of the model by attributing relevance scores to every input dimension based on the model's prediction. It has been shown that several of these methods are special cases of the more general game-theoretic concept of Shapley values \citep{lundberg2017unified}. Layer-wise relevance propagation \citep[LRP;][]{bach2015pixel} is among these methods and propagates relevance backward through the network from one layer to another. The basic LRP approach was subsequently developed and extended to further network architectures \citep{binder2016layer, arras2017explaining, schnake2021higher, ali2022xai, montavon2019layer}. It has gained increased popularity since then and is widely used to explain a model's prediction.
\begin{figure}[H]
	\centering
	\includegraphics[scale=0.3]{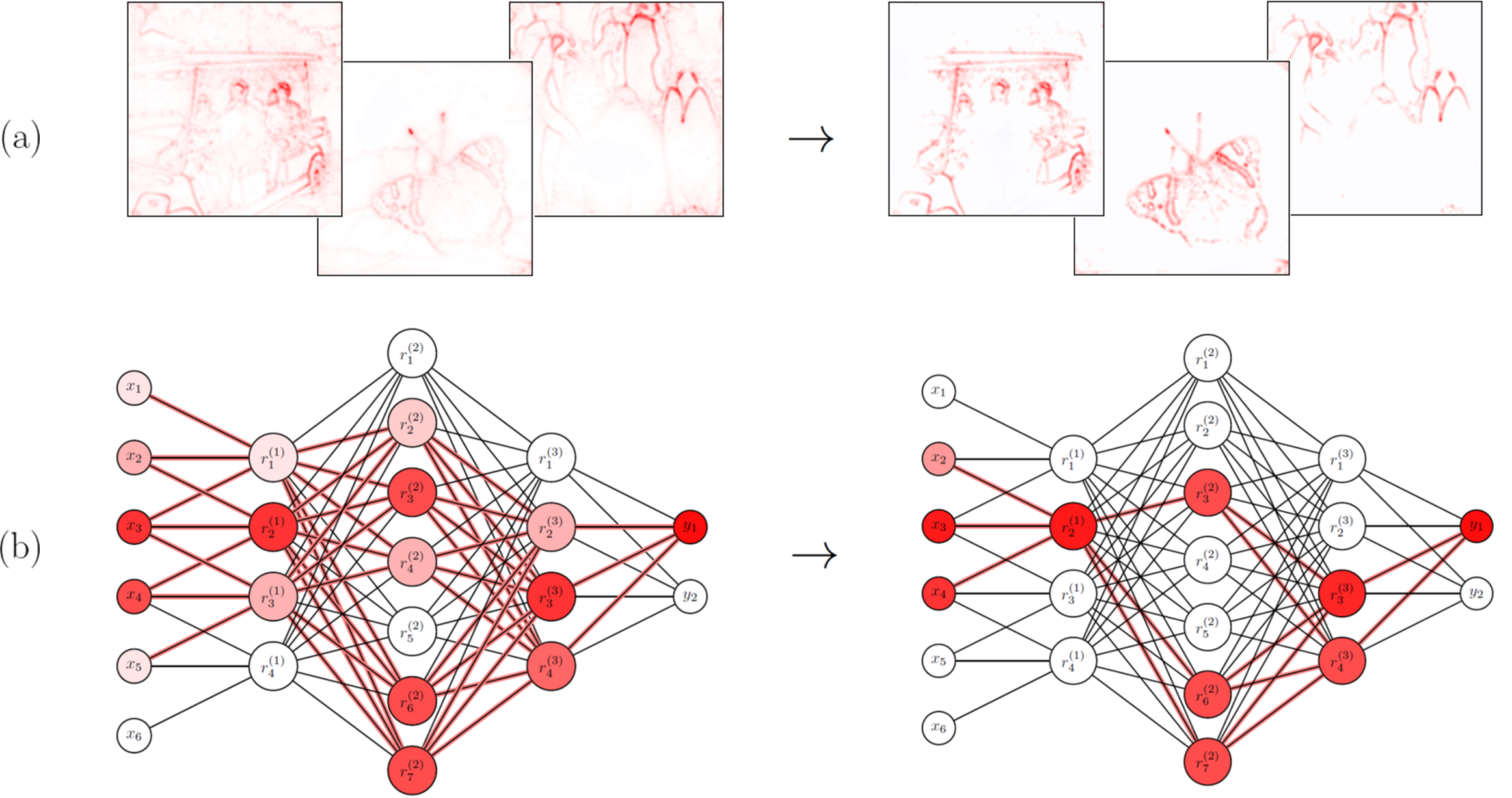}
	\caption{Illustration of sparsified explanations using our pruned layer-wise relevance propagation (PLRP): (a) Heatmaps of explanations for image classifications obtained through LRP (left) and their corresponding sparsified versions through PLRP (right). These sparser relevance attributions are more distinct explanations of the images and can help to identify and interpret the most important features. (b) Corresponding (exemplary) DNNs show how neurons with low relevance are removed, leading to sparser relevance attribution in every layer and fewer paths through which relevance is propagated. These sparser relevance attributions in the intermediate layers potentially allow to better understand latent factors of the model.}
    \label{fig1}
\end{figure}
For image classification, a human evaluation and interpretation of the LRP explanation can be done qualitatively by a suitable visualization such as heatmaps. The observer can identify the object to be detected and decide whether the corresponding relevance attribution split across the input dimensions matches this object. However, for high-dimensional data with unknown ground truth, this might not be feasible anymore. For example, this is often the case for genome sequences. Besides the high dimensionality, noisy relevance attributions can make it difficult to identify and interpret the main drivers for the model's prediction in such data and hence, diminish the meaningfulness of the explanation. As a solution, we propose a modification of LRP that generates sparser explanations by extending the idea of filtering or pruning the relevance propagation \citep{montavon2018methods}. We call our approach \textit{pruned layer-wise relevance propagation} (PLRP). Sparsification of the explanation might be desirable in the sense that it reduces noise and the number of features with non-zero relevance, i.e., highlights only the most important features. Furthermore, our method is able to regulate the degree of sparsity, which can be beneficial for feature selection or the identification of concepts learned by the network in deeper layers. While our approach is specific to LRP, the general idea of enforcing sparsity is not. In fact, a similar modification might be applicable to other local explanation methods. The basic intuition of our method is illustrated in  Figure \ref{fig1}(a) via three heatmaps of images obtained through LRP (left) and their corresponding sparsified versions through PLRP (right). Further, Figure \ref{fig1}(b) shows how the relevance propagation in our proposed method PLRP is pruned within a DNN by removing less relevant neurons. This idea of maintaining only the most important features is inspired by $L1$-regularization. However, integrating  $L1$-regularization into LRP is not directly applicable. This is because the map representing the relevance propagation from one layer to another, defined by LRP, is input-specific. Hence, there is no single consistent map for all data points. Approximating all inputs with the same model would be a strong simplification of the original explanation method. It would also disregard the fact that different inputs might activate different neurons in intermediate layers. Altogether, this motivates our contribution of PLRP.

\section{Related Work}\label{sec_related_work}

Since LRP was introduced by \cite{bach2015pixel}, several modifications of it have been proposed. Some work focuses on creating class-discriminative explanations by not only creating relevance attributions for the target class.  \cite{gu2019understanding} and \cite{iwana2019explaining} propose modifications where, also for non-target classes, the negative prediction score is propagated backward. Hence, the final explanation is the difference between target and non-target classes. Similarly, \cite{montavon2019layer} point out that instead of the prediction score, one could also propagate the log-probability ratios to achieve class-discriminative explanations. However, \cite{jung2021explaining} show that some of these modifications exhibit a so-called erasing object problem. This means when considering differences in explanations, some relevance scores might cancel each other out and thereby potentially falsely erase relevant features from the explanation. To overcome this issue, they introduce a modification of LRP by constraining the relevance propagation to neurons with positive gradient. Hence, when propagating backward, the relevance of some neurons is set to zero. \cite{montavon2018methods} also present this idea of filtering the propagation through certain neurons. This is also used by \cite{achtibat2022towards} to create more human-understandable explanations, assuming there is some understanding of the representation of data in the latent model space, i.e., the roles of the learned latent factors. They first identify these factors globally for the model and then restrict the propagation of relevance through them locally per input sample. \cite{chormai2022disentangled} propose a numerical approach to identify relevant subspaces of the latent space for models with unknown roles of latent factors. These subspaces allow to create disentangled explanations that decompose the explanation into different components, which can then be visualized and interpreted. 

\cite{adebayo2018sanity} and \cite{sixt2020explanations} conclude that some explanation methods, such as LRP, might be insensitive to changes in the model's parameters and hence, question their reliability. However, in a later study, \cite{arras2022clevr} cannot confirm these results. In fact, in their evaluation, LRP is among the best-performing explanation methods. The authors argue the insensitivity might be more likely driven by non-class-discriminative explanations or only asymptotically relevant.

Although related to model pruning \citep{zhu2017prune}, our approach and goal of pruning relevance propagation differ conceptually from it. Instead of pruning the underlying model, we aim to sparsify the individual explanations, which are local approximations of the model. Hence, we prune different neurons depending on the input features, thereby enabling a local and input-specific pruning. In contrast, when \cite{gupta2021obtaining} and \cite{yeom2021pruning} applied LRP for model pruning, they used the relevance attribution aggregated over multiple inputs to prune the model globally. The authors show that the number of parameters can be reduced drastically, while still maintaining or even improving the predictive performance of the model.

\section{Pruned Layer-Wise Relevance Propagation (PLRP)}\label{sec_dlrp}

We propose to extend the idea of filtering the propagation of relevance through certain neurons \citep{montavon2018methods} by using the relevance itself as filter criterion. To the best of our knowledge, none of the previous works focused on creating sparse explanations in the sense of locally reducing the input features and latent factors to the most important ones and concentrating relevance mass there. As the explanations are input-specific, a local pruning of relevance propagation is more appropriate for our goal than a global pruning of the model. This motivates our \textit{pruned layer-wise relevance propagation} (PLRP).

\subsection{Setting}
Let $f:\R^d\rightarrow\R^C, \bx=(x_1,\ldots, x_d)^\top \mapsto f(\bx)=(f_1(\bx), \ldots, f_C(\bx))^\top$ be the pre-softmax prediction function of a trained DNN for a classification task with $C$ classes, which we also refer to as \textit{model}. For our setting, we are only interested in the prediction score of the winning class, denoted by $f_{c^*}(\bx)$. For a given model, an explanation method is a function $e: \R^d\rightarrow\R^d: \bx\mapsto e(\bx)=(r_1, ..., r_d)^\top$ that assigns a relevance score $r_i$ to every input dimension $i=1,\ldots, d$. We denote by $W_l\in\R^{m\times n}$ the weight matrix of the $l$-th layer of the model and by $w_{jk}^{(l)}$ the weight of the edge connecting the $j$-th neuron in layer $l-1$ with the $k$-th neuron in layer $l$. The corresponding activations are denoted by $a_j^{(l-1)}$ and $a_k^{(l)}$, respectively. Hence, we can write $\ba^{(l)}=\big(W^\top_l\ba^{(l-1)}\big)^+$, where $(\cdot)^+$ refers to the entry-wise application of the ReLU function. Since for our purposes we consider only one layer at a time, we will drop the layer-specific superscripts for readability and simply write $w_{jk}$. Furthermore, let $(\cdot)^-$ be the function that only keeps the absolute value of the negative part and is zero otherwise.

 Finally, for the relevance attribution, we refer to the relevance of the $k$-th neuron in layer $l$ as $r_k^{(l)}$ and write $\br^{(l)}=(r_1^{(l)}, \ldots,  r_n^{(l)})^\top$, in particular $\br^{(0)} = \br = e(\bx)$. For the last layer $L$ by definition, we have $r_{c^*}^{(L)}:=f_{c^*}(\bx)$ and $r_i^{(L)}:=0, i\neq c^*$.

As described earlier, LRP propagates the relevance backward from one layer to its predecessor, starting with the prediction score of the output layer $f_{c^*}(\bx)$. The total relevance mass is preserved in every layer by the so-called conservation property \citep{bach2015pixel}. The relevance propagation from layer $l$ to $l-1$ can be expressed by a linear map \citep{sixt2020explanations}, i.e., a multiplication with the matrix $M_l: \R^n\rightarrow\R^m: \br^{(l)}\mapsto\br^{(l-1)}=M_l\br^{(l)}$. Thereby, the matrix entry in the $i$-th row and $j$-th column represents the proportion of relevance of neuron $j$ in layer $l$ that is allocated to neuron $i$ in layer $l-1$. As for the weights, we will disregard the layer subscript and only write $M$. There are different LRP rules for how these proportions are calculated \citep{montavon2018methods}. Specifically, for the LRP-0 rule, the matrix $M$ is given by
\begin{align}\label{eq_matrix_M}
    M=
    \begin{bmatrix}
    \frac{a_jw_{jk}}{\sum_{j=1}^m a_jw_{jk}}
    \end{bmatrix}_{j=1, ..., m, k=1, ..., n} \,.
\end{align}
Note that by definition, the columns of $M$ are normalized corresponding to redistributing the whole relevance of a neuron. For choosing a rule that yields measurably  appropriate relevance attributions, we follow the  composite strategy of \cite{kohlbrenner2020towards} that combines various rules for different layer types.

\subsection{Pruned LRP}

 The goal of \textit{pruned layer-wise relevance propagation} (PLRP) is to create sparser explanations by reducing noise and simultaneously increasing the attribution to the most relevant features. We do this by pruning the explanation directly in every layer. This consists of two steps: first, determining the neurons to be pruned, and second, redistributing their relevance mass among the remaining neurons. The criterion for pruning the neurons is the relevance itself, i.e., the neurons with the lowest relevance scores get pruned. Hence, first, we conduct a preliminary unpruned relevance attribution to layer $l$, prune it, and then after redistribution, use it as input for the attribution to layer $l-1$. As it is unclear in advance how the relevance is distributed among the neurons, we prune a proportion of the total relevance mass instead of a proportion of the neurons.

By redistributing the pruned relevance mass among the remaining neurons, we maintain the relevance conservation of LRP. As we are potentially interested in strong negative contributions, we consider the positive and negative relevance separately, i.e., we aim to prune neurons with small relevance in absolute terms. Hence, in the following by $\br^{(l)}$ we either mean $\big(\br^{(l)}\big)^+$ or $\big(\br^{(l)}\big)^-$. Note that the proportion of pruned mass can be different for the positive and negative parts of the relevance vector. We do not prune the relevance propagation in the last step, i.e., the first layer is left unpruned. This would correspond to a simple thresholding of the input because it is not further propagated through further layers and thus is neither desired nor meaningful.

\subsubsection{Pruning Relevance}

We denote the unpruned relevance vector by $\widetilde{\br}^{(l-1)}$. Then for a pre-defined proportion $p_l\in[0,1)$ of the total relevance mass that is supposed to be pruned, we determine the corresponding threshold $\theta_l\in\mathbb{R}$, such that the relevance scores below $\theta_l$ sum up to $p_l$ of the total relevance mass. Hence, for (unpruned) ordered relevance scores $\widetilde{r}_{s_1}^{(l-1)}\le \widetilde{r}_{s_2}^{(l-1)}\le \ldots \le \widetilde{r}_{s_n}^{(l-1)}$, we set
\begin{align}\label{eq_threshold_theta}
    \theta_l:=\max_{1\le i^*\le n} \Bigg\{\widetilde{r}_{s_{i^*}}^{(l-1)} \,\Bigg|\, \sum_{i=1}^{i^*} \widetilde{r}_{s_i}^{(l-1)}\le p_l\sum_{i=1}^n \widetilde{r}_i^{(l-1)}\Bigg\} \,.
\end{align}
If $p_l$ leads to an empty set in \eqref{eq_threshold_theta}, we set $\theta_l:=0$. Hence, we obtain the desired relevance pruning as
\begin{align*}
\left\|\boldsymbol{1}_{\left\{\widetilde{\br}^{(l-1)}>\theta_l\right\}}\odot\widetilde{\br}^{(l-1)}\right\|_1 \ge (1-p_l)\left\|\widetilde{\br}^{(l-1)}\right\|_1 \,,
\end{align*}
where $\odot$ denotes the element-wise multiplication and $\boldsymbol{1}_{\{\,\boldsymbol{\cdot}\,>\theta_l\}}$ is the element-wise application of the indicator function. Hence, we only keep the entries where the neuron's relevance lies above the threshold $\theta_l$. 
Every $p_l$ can be seen as a hyperparameter that could be tuned. However, for dense networks, this can easily become computationally very expensive. Further, as the relevance attribution is input-specific, the optimal parameter might differ significantly for different samples. Therefore, we consider an additional approach that determines $\theta_l$ and hence, $p_l$ after the relevance propagation in every layer. As we are interested in sparse relevance vectors, we consider the so-called \textit{sparsity gain}. Thereby, the additional degree of sparsity with respect to pruned relevance is calculated. The idea is that neurons with small relevance yield a high gain, while the gain decreases for neurons that concentrate mass as more mass must be pruned for an additional degree of sparsity. Since the total relevance may differ for different inputs, we consider the relative pruning, i.e., the proportion $p_l$. Therefore, consider a real-valued vector $v_0$. We iteratively sparsify $v_0$ by setting some entries to zero. Let $v'$ denote the vector after setting some entries in $v_0$ to zero and $v''$ the vector after further entries in $v'$ were set to zero. Then for some sparsity measure $s$ that increases with respect to the degree of sparsity, the sparsity gain for setting $v'$ to $v''$ is defined as
\begin{align}
    \Delta_{SG} := \frac{\Delta s}{\Delta p_l} = \|v_0\|_1\frac{s(v'')-s(v')}{\|v''\|_1-\|v'\|_1}\,.
\end{align}
When setting entries in $v_0$ iteratively to zero based on their increasing value, then the sparsity gain is monotonically decreasing. Hence, a minimal allowed sparsity gain $\Delta_{SG}$ determines a proportion of pruned mass $p_l$ and thus a threshold $\theta_l$. Therefore, the approach of sparsity gain makes PLRP independent of choosing fixed proportions $p_l$ in advance. Instead, every $p_l$ is determined during the process of pruning. This procedure is related to considering a minimal or maximal change of a function, i.e., its derivative. Nevertheless, the choice of a minimal sparsity gain is necessary. A natural choice would be 1, i.e., when an additional degree of sparsity becomes more expensive than one unit of pruned mass.

\subsubsection{Redistributing Pruned Relevance}

The pruned relevance is redistributed among the remaining neurons with non-zero relevance. We consider two possible approaches. The first one is based on the relative proportion of the neuron's relevance after pruning. This is equivalent to a rescaling by a factor $\lambda$, such that the total relevance mass sums up again to the value before pruning. We refer to this procedure as \textit{PLRP}-$\lambda$. For this, define
\begin{align*}\label{eq_PLRP_lambda}
    \br^{(l-1)} := \lambda_l\,\boldsymbol{1}_{\big\{\widetilde{\br}^{(l-1)}>\theta_l\big\}} \widetilde{\br}^{(l-1)} \,,
\end{align*}
with
\begin{align*}
    \lambda_l = \left\|\widetilde{\br}^{(l-1)}\right\|_1 \,\Big/\,\left\|\boldsymbol{1}_{\left\{\widetilde{\br}^{(l-1)}>\theta_l\right\}}\odot\widetilde{\br}^{(l-1)}\right\|_1\approx\frac{1}{1-p_l} \,.
\end{align*}
For the second approach, denoted by \textit{PLRP}-$M$, we make use of the fact that LRP assigns zero relevance to neurons with zero activation, i.e., $a_j^{(l)}=0 \Rightarrow r_j^{(l)}=0$. Therefore, (re-)applying LRP with more zero activations leads to sparser relevance attribution without the necessity of rescaling the relevance mass. We do this by modifying the matrix $M$ expressing the relevance propagation from Equation \eqref{eq_matrix_M}. Given the unpruned relevance vector $\widetilde{\br}^{(l-1)}$, we define the modified activations by
\begin{align}
    \widehat{\ba}^{(l-1)} &:= \boldsymbol{1}_{\left\{\widetilde{\br}^{(l-1)}>\theta_l\right\}}\odot \ba^{(l-1)}
\end{align}
and hence, the modified relevance attribution as
\begin{align}\label{eq_matrix_Mhat}
    \widehat{M}_l &:=
        \begin{bmatrix}
            \frac{\widehat{a}_j^{(l-1)}w_{jk}^{(l)}}{\sum_{j=1}^m \widehat{a}_j^{(l-1)}w_{jk}^{(l)}}
        \end{bmatrix}_{j=1, ..., m, k=1, \ldots, n} \\
        \br^{(l-1)} &:= \widehat{M}_l\br^{(l)} \,.
\end{align}
Hence, by attributing relevance according to $\widehat{M}_l$, we ensure that neurons receive zero relevance that would have been assigned a relevance mass below the threshold $\theta_l$ of Equation \eqref{eq_threshold_theta} otherwise. As $\widehat{M}_l$ is also a matrix in the form of Equation \eqref{eq_matrix_M}, only with more zero activations, the pruned relevance gets implicitly redistributed in the manner of LRP.

\section{Evaluation}\label{sec_evaluation}

We evaluate the two variants of PLRP in the context of classification of two different data types: images and genome sequences. For the quantitative evaluation, we use multiple metrics covering different aspects of the explanation method following the categorization of \cite{hedstrom2023quantus}. Further, we evaluate our methods qualitatively by considering heatmaps and sequence logo plots of the relevance attributions. 

\subsection{Evaluation Metrics}

\textit{Sparsity} and \textit{localization} are intended to measure the efficacy of our method, while \textit{robustness} and \textit{faithfulness} serve as sanity checks for its trustworthiness. Note that since these metrics consider different aspects, they do not necessarily align. $\uparrow$ and $\downarrow$ hint whether a higher or lower value indicate better outcomes.
\vspace{-0.4cm}
\paragraph{Sparsity}
As we are interested in sparse explanations that concentrate the relevance mass at few important features, we evaluate the degree of sparsity or complexity by the Gini Index ($\uparrow$) and entropy ($\downarrow$) of the relevance attribution as proposed by \cite{chalasani2020concise} and \cite{bhatt2020evaluating}. Thereby, a higher Gini Index or lower entropy, respectively, indicate that there are more zero entries and more mass is concentrated at fewer entries.
\vspace{-0.4cm}
\paragraph{Localization}
Localization measures how accurately the relevance is attributed to a pre-specified region. Hence, for this metric, a known ground truth is needed. Therefore, we calculate the \textit{Relevance Mass Accuracy} \citep[RMA, $\uparrow$;][]{arras2022clevr}, which measures how much relevance mass out of the total mass is allocated to the ground truth.
\vspace{-0.4cm}
\paragraph{Robustness}
For explanation methods, robustness refers to the idea that similar inputs should produce similar explanations. Hence, it can be seen as a sensitivity proxy that measures how much the explanation changes for a small shift in the input. \cite{alvarez2018robustness} propose to estimate the local Lipschitz constant ($\downarrow$), i.e., for $\varepsilon>0$ and some input $\bx\in\R^d$, the aim is to find a constant $L=L(\bx)$ such that for all $\bx'\in B_\varepsilon(\bx)=\{\bx' :\|\bx-\bx'\|<\varepsilon\}$ it holds $\|e(\bx)-e(\bx')\|<L\|\bx-\bx'\|$. Hence, $L$ is the steepest possible ascent of $e$ in a small neighborhood of $\bx$. As calculating the gradient of the explanation method might be very expensive, or $e$ might not even be differentiable, $L$ is estimated numerically by sampling from $B_\varepsilon(\bx)$.
\vspace{-0.4cm}
\paragraph{Faithfulness}
Faithfulness measures whether the explanation method indeed uncovers the predictive behavior of the underlying model. This means that the highest relevance scores correspond to the most decisive input features. The idea is to perturb the input based on the relevance attribution and then measure how quickly the predictive score drops. This procedure is also referred to as \textit{pixel-flipping} \citep{bach2015pixel} or \textit{selectivity} \citep{samek2016evaluating}. Hence, we have a perturbation function that takes the model input $\bx$ and corresponding relevance attribution $\br=e(\bx)$ and outputs a sequence of perturbed versions of $\bx$ based on $\br$, i.e., $(\bx, \br)\mapsto(\bx_1',\ldots, \bx_s')$ where $s$ denotes the number of perturbation steps. As the actual value of prediction scores can differ from one class to another, we normalize the scores and measure the relative drop. Further, to compare the results across different pruning parameters $p_l$, we also consider the area under the curve (AUC, $\downarrow$). Note that a quicker decrease in the prediction score corresponds to a lower AUC and hence, to a better explanation with respect to this metric. As faithfulness might be sensitive to the choice of perturbation, e.g., size of perturbed area or type of noise to be added, this metric should be considered cautiously. This holds especially when comparing small differences in this metric. Nevertheless, it can serve as a proxy for the general trend in how faithful a method is.

\subsection{Experiments}

\subsubsection{Image Classification}

\paragraph{Model Specification}
We perform our evaluation on the ImageNet dataset \citep{russakovsky2015imagenet} and the Extended Complex Scene Saliency Dataset (ECSSD) \citep{shi2015hierarchical}. Thereby, we consider two established model architectures: VGG-16 \citep{simonyan2014very} and ResNet-50 \citep{he2016deep}, both convolutional neural networks (CNN). The ground truth masks of ECCSD are more granular than the bounding boxes of ImageNet, which contain more pixels than the actual ground truth object. Therefore, ECCSD is more suitable for measuring localization. However, for comparison and completeness, we measure sparsity, robustness, and faithfulness also on the widely used ImageNet dataset. Thereby, we use a sample of 3,900 images of its validation set covering all classes. As we must draw multiple samples to calculate the Lipschitz estimate for a single input for robustness analyses, we restrict the number of input samples to 10. The evaluation includes both of our approaches, PLRP-$\lambda$ and PLRP-$M$, with and without sparsity gain. Thereby, we examine multiple values for the parameter $p\in[0,1)$, i.e., the proportion of pruned relevance per layer. Specifically, we consider a range from 0 to 0.95 in increments of 0.05.

\begin{figure}[H]
	\centering
	\includegraphics[scale=0.25]{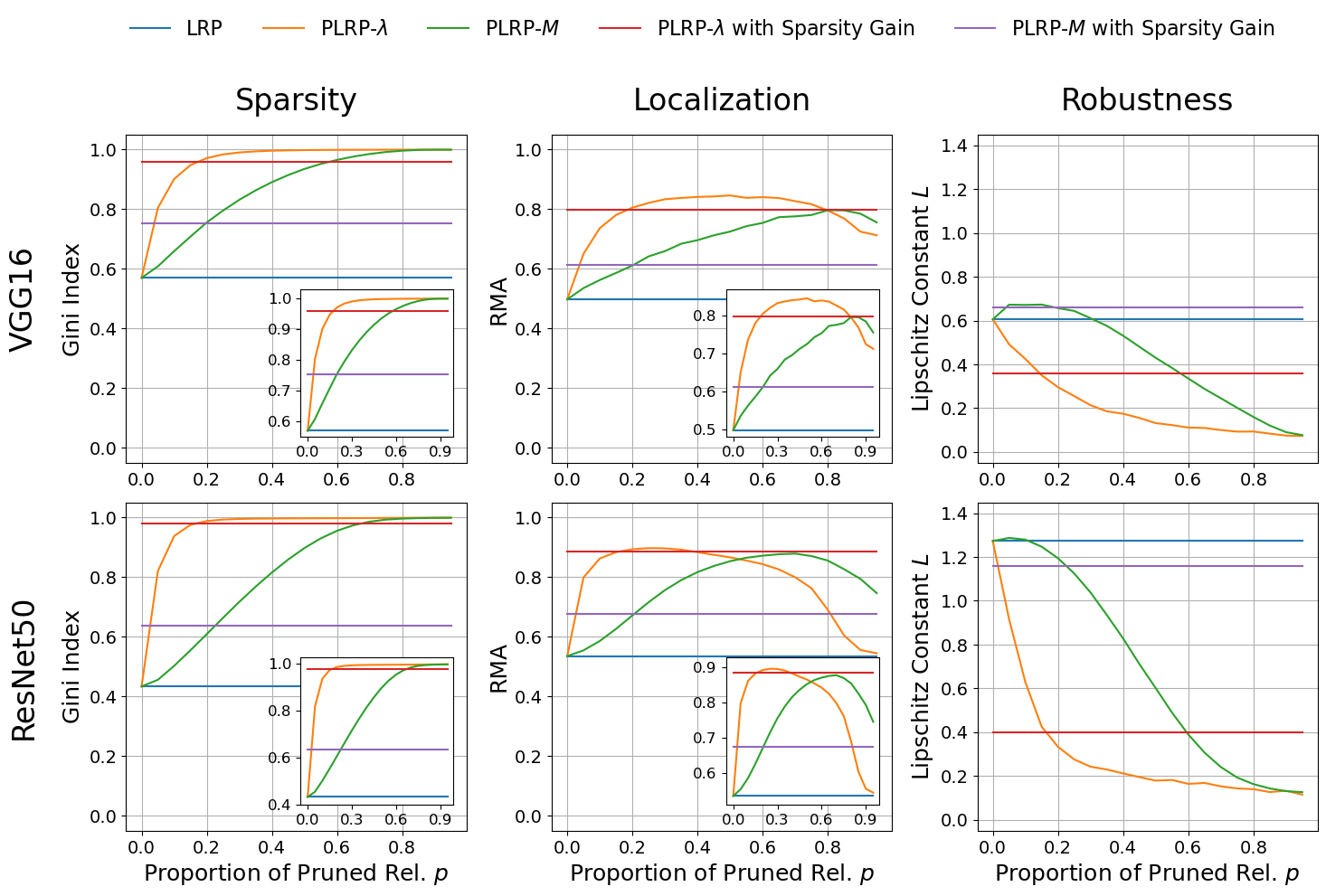}
	\caption{Results for ECSSD for metrics sparsity, localization, and robustness for different proportions of pruned relevance mass $p$ ranging from 0 to 0.95 in increments of 0.05 for models VGG16 and ResNet50. For sparsity and localization, the y-axis covers the whole output domain of $[0,1]$. The zoom plots focus on the actual covered output domain for better comparison of the methods.}
    \label{fig_image_results}
\end{figure}

Our implementation of PLRP\footnote[1]{Code available at \url{https://gitlab.com/dacs-hpi/plrp}} is based on the \textit{Zennit} package \citep{anders2021software}. For the evaluation, we use the \textit{Quantus} package \citep{hedstrom2023quantus} that implements several evaluation metrics. For evaluating faithfulness, we use a feature of the \textit{iNNvestigate} package \citep{alber2019innvestigate} that perturbs the input $\bx$ based on its explanation $e(\bx)$.
\vspace{-0.4cm}
\paragraph{Results}
We observe that our approach creates sparser explanations, while maintaining the most relevant features. Figure \ref{fig_image_results} illustrates that the Gini Index increases drastically for all our approaches compared to the LRP baseline. Simultaneously, the RMA increases, indicating that after pruning and redistributing a higher proportion of relevance falls within the ground truth mask. In both metrics, we observe a steep increase already for small proportions ($<0.2$) of pruned relevance $p$. For both models, VGG16 and ResNet50, the simpler approach PLRP-$\lambda$ with and without sparsity gain leads to better results compared to PLRP-$M$ and the LRP baseline. We also find that the RMA starts decreasing again for higher $p$. This is not surprising as, at some point, the most decisive features might also get pruned. These findings are in line with what we observe qualitatively by visualization (see Figure \ref{fig_heatmaps}), i.e., PLRP-$\lambda$ produces sparser explanations than PLRP-$M$ for the same pruning parameter $p$. For higher $p$, also relevance within the ground truth mask gets pruned. The entropy shows completely analogous results to the Gini Index and hence, is not displayed here.

Furthermore, PLRP-$\lambda$ is always more robust than the LRP baseline, whereas PLRP-$M$ is less robust than the baseline for only a few parameterizations. Thus, we conclude that the explanations of our approach can be seen as reliable and stable.

\begin{figure}[H]
	\centering
	\includegraphics[scale=0.25]{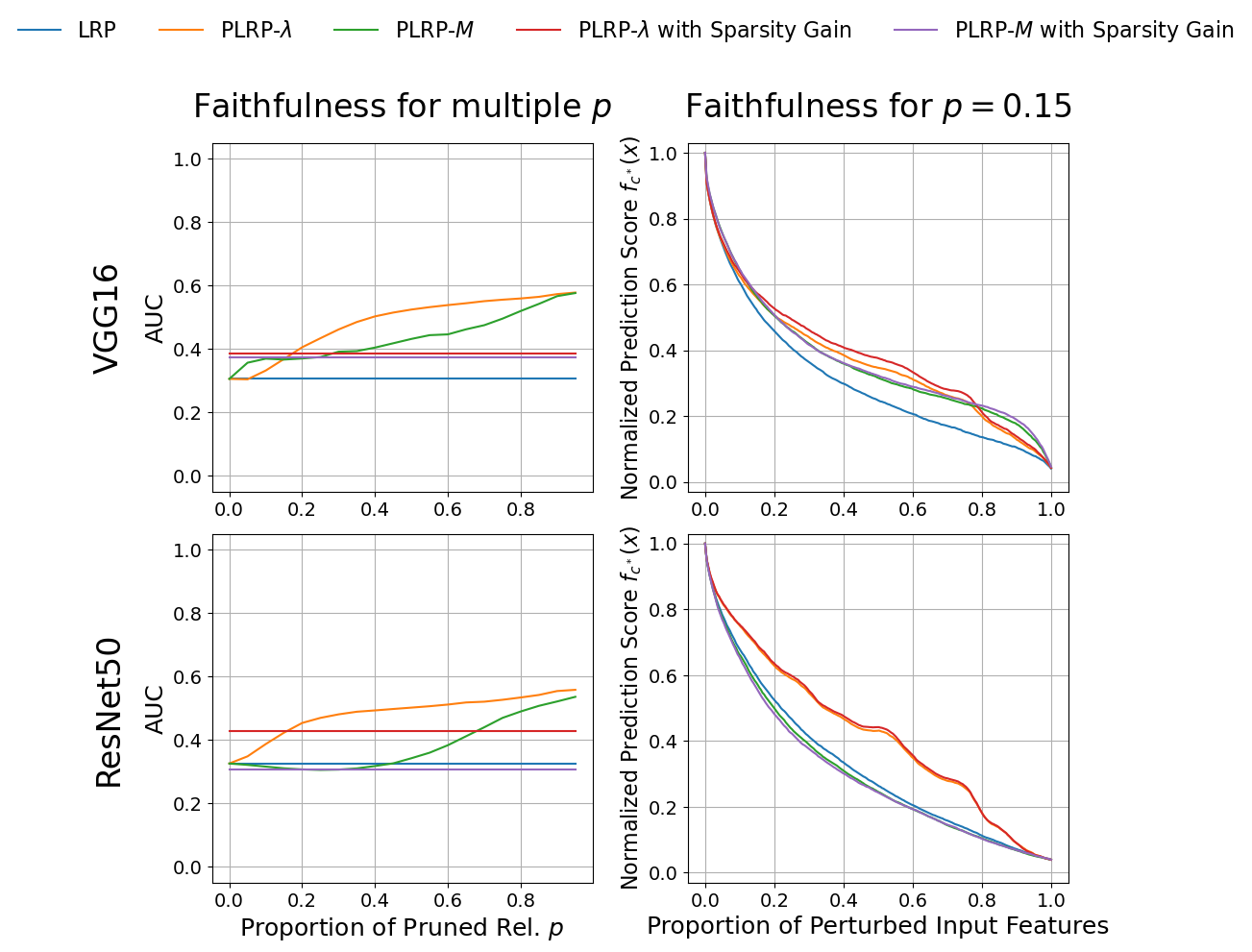}
	\caption{Results for ECSSD for faithfulness for different proportions of pruned relevance mass $p$ ranging from 0 to 0.95 in increments of 0.05 (left) and for an exemplary parameterization $p=0.15$ (right) for models VGG16 and ResNet50. The AUC differs only slightly compared to the LRP baseline (left). For higher $p$, the difference becomes larger as more features are pruned. For the exemplary parameterization, the prediction score $f_{c^*}(\bx)$ drops similarly steeply as for the LRP baseline for the features with the highest relevance that are perturbed first (right). The difference in AUC is driven by the less important features that are perturbed later.}
    \label{fig_image_results_faithfulness}
\end{figure}

\begin{figure}[H]
	\centering
	\includegraphics[scale=0.24]{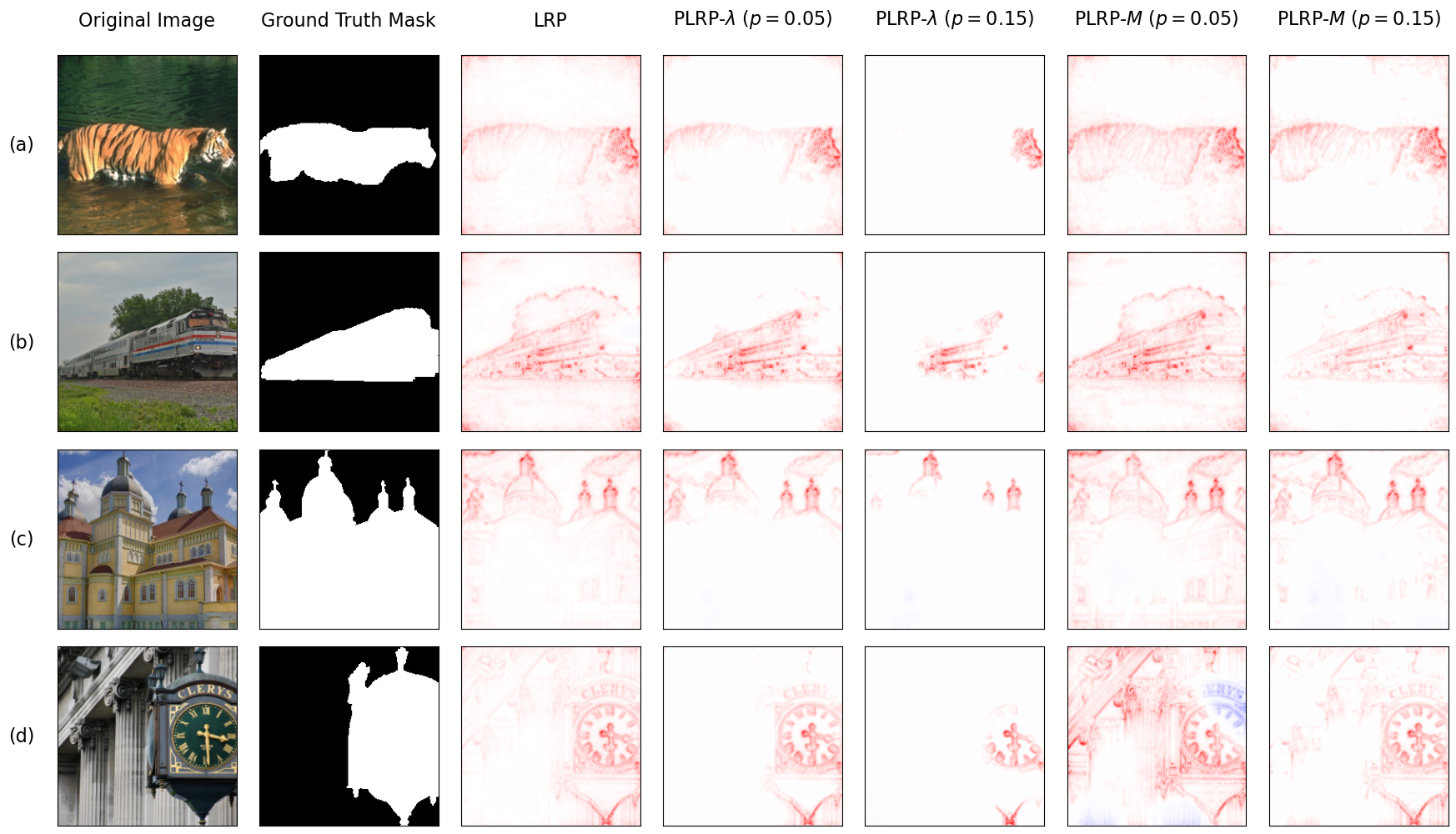}
	\caption{Illustration of relevance attribution via heatmaps for LRP and PLRP with different parameterizations for the VGG16 model. Red pixels indicate positive relevance, while blue are negative. PLRP-$\lambda$ and PLRP-$M$ produce sparser explanations than the baseline LRP, thereby concentrating relevance within the ground truth mask. For the same pruning parameter $p$, PLRP$\lambda$ prunes more features than PLRP-$M$. For higher $p$, both approaches prune relevance within the ground truth mask.}
    \label{fig_heatmaps}
\end{figure}

With respect to faithfulness, except for one approach (PLRP-$M$ with and without sparsity gain and $p<0.3$), we observe slightly worse results than the baseline (see Figure \ref{fig_image_results_faithfulness}). Note that better results correspond to a lower AUC as we are interested in a quick drop in the prediction score. However, we observe that the higher AUC is mainly driven by less important features, i.e., we have a less steep decrease in the prediction score for later perturbed and hence, less important features. This is in line with our goal of maintaining only the most important features. In fact, for the features with the highest relevance, the prediction score drops similarly steeply as for the baseline (see Figure \ref{fig_image_results_faithfulness} (right)). Finally, the relative decline in faithfulness is small compared to the relative gain in sparsity and localization, especially for small $p$.

Overall, PLRP-$\lambda$ is able to increase sparsity quicker and concentrate more relevance within the ground truth mask than PLRP-$M$, i.e.,  there is a steeper increase for the same proportion of pruned relevance. It appears that redistributing the pruned relevance in the manner of LRP as done by PLRP-$M$ slows down the concentration of relevance. This is also in line with PLRP-$M$ sticking closer to the baseline with respect to faithfulness.
Another driver for the better results of PLRP-$\lambda$ over PLRP-$M$ with respect to sparsity, localization, and robustness appears to be that PLRP-$M$ can lead to a sign flipping of a neuron's relevance (see negative relevance for PLRP-$M$ in Figure \ref{fig_heatmaps}(d)). PLRP-$M$ changes the proportions of how much relevance is attributed from one neuron to another (see Equation \eqref{eq_matrix_Mhat}). Since this might increase the proportion of incoming negative relevance or decrease it for incoming positive relevance, respectively, the sign of the total relevance a neuron receives might switch. Furthermore, the sign of the proportions themselves might change. PLRP-$\lambda$, on the other hand, only rescales the relevance by a positive factor and hence, preserves the initial sign of the relevance score.

The quantitative results for the ImageNet dataset for sparsity, robustness, and faithfulness are analogous to the results for ECSSD. Hence, the experiments on ImageNet confirm the results above, thereby providing further confidence in the findings. They can be found in the supplementary material in section A.

\subsubsection{Genomics}

\begin{figure}[H]
	\centering
	\includegraphics[scale=0.48]{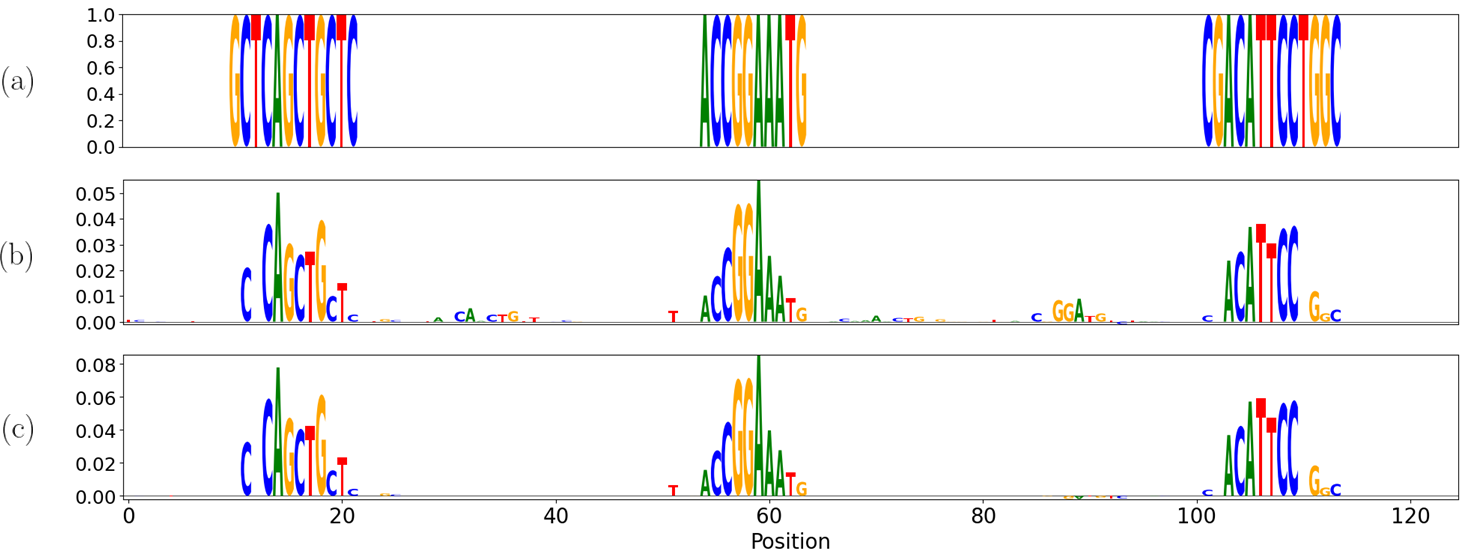}
	\caption{Illustration of pruned explanations: Logo plots of (a) ground truth mask, (b) relevance scores from LRP, and (c) PLRP-$\lambda$ with $p=0.25$ for the model’s prediction. PLRP reduces noise and creates a sparser explanation compared to the LRP baseline. Thereby, PLRP maintains features that lie within the ground truth even if they have less relevance than the noise at other irrelevant features. Hence, it indeed differs from only applying a threshold.}
    \label{fig_logoplot1}
\end{figure}

\paragraph{Model Specification}
Next, we apply PLRP to explain predictions of CNNs trained on synthetic genome sequences data with known ground truth \citep{lemanczyk2024motif}. Specifically, the input consists of strings of length 250 from the alphabet $\{A,C,G,T\}$ representing the four base pairs in DNA. The sequences are one-hot-encoded. Hence, our input domain is $\{0,1\}^{4\times250}$. Thereby, we consider two different model architectures: one with only four filters and another one with 32. As we operate on a discrete domain, slight changes in the input or perturbations are not well-defined. Therefore, we only calculate the metrics for sparsity and localization.
\vspace{-0.4cm}
\paragraph{Results}
Similar to the task of image classification, both PLRP variants produce sparser explanations compared to the LRP baseline. We observe qualitatively (see Figure \ref{fig_logoplot1}) that noise is reduced and relevance is concentrated at the ground truth patterns in the genome sequence (motifs). Similar to the application to image classification, we find that, in general, the RMA increases for an increasing proportion of pruned relevance $p$. Hence, the more we prune and redistribute, the more relevance lies within the ground truth mask. Simultaneously, sparsity rises, while complexity is reduced as indicated by an increasing Gini Index.

Figure \ref{fig_logoplot1} also shows that this noise reduction indeed differs from simple thresholding, as it maintains relevance scores within the ground truth mask that are smaller than the noise signal at other irrelevant features.

The results of PLRP-$\lambda$ and PLRP-$M$ only differ slightly. This might be due shallower networks used here compared to the image classification. The fewer layers there are, the less both variants can diverge from each other. The quantitative results for the genomics application can be found in the supplementary material in section B.

\section{Conclusion}\label{sec_conclusion}

In this work, we introduced PLRP, a modification of LRP that enforces sparsity directly by locally pruning the relevance propagation for the different layers. We presented two approaches to redistribute the pruned relevance mass so that the conservation property is preserved: the simpler PLRP-$\lambda$ and the more complex PLRP-$M$, which is more in line with original LRP methodology. The evaluation on the ECSSD and ImageNet datasets shows that we indeed obtain sparser explanations than the LRP baseline. The pruning leads to only a slight decrease in faithfulness, mainly driven by the least important features. By measuring localization, we show that while becoming sparser, more relevance is attributed to features within the ground truth mask. This demonstrates the efficacy of PLRP compared to LRP. In fact, both our approaches lead to noise reduction and concentrate relevance at the most important features. Qualitative evaluation by visualization supports this claim. Furthermore, our approach is similar and, depending on the parameterization, even more robust to small changes in the input than the LRP baseline. For the genomics application, we observe similar effects. We obtain sparser explanations with higher concentration of relevance on the most important features in the genome sequence, aiding in the interpretation of the model outputs.
Overall, PLRP-$\lambda$ produces better results for our goal of sparsity and relevance concentration than PLRP-$M$. This is partially driven by the effect of relevance sign flipping. However, it should be investigated whether this effect can be reduced by further modification. Additionally, how to define and find an optimal parameterization for PRLP remain open questions. This might heavily depend on the underlying model but also on the ``optimal'' degree of sparsity for a specific task. Further, as PLRP not only prunes the relevance attribution for the input features but also for neurons in intermediate layers, it could be used to study what concepts a model learned in deeper layers. Finally, as the general idea of enforcing sparsity when attributing relevance is not specific to LRP, it can also be applied to other explanation methods.
%% ====================================================================================================================================================================================== %%
%% ====================================================================================================================================================================================== %%
\bibliographystyle{apalike}
\bibliography{bib}

\appendix
\newpage
\section*{Appendix}

\subsection*{A}
\begin{figure}[H]
	\centering
	\includegraphics[scale=0.25]{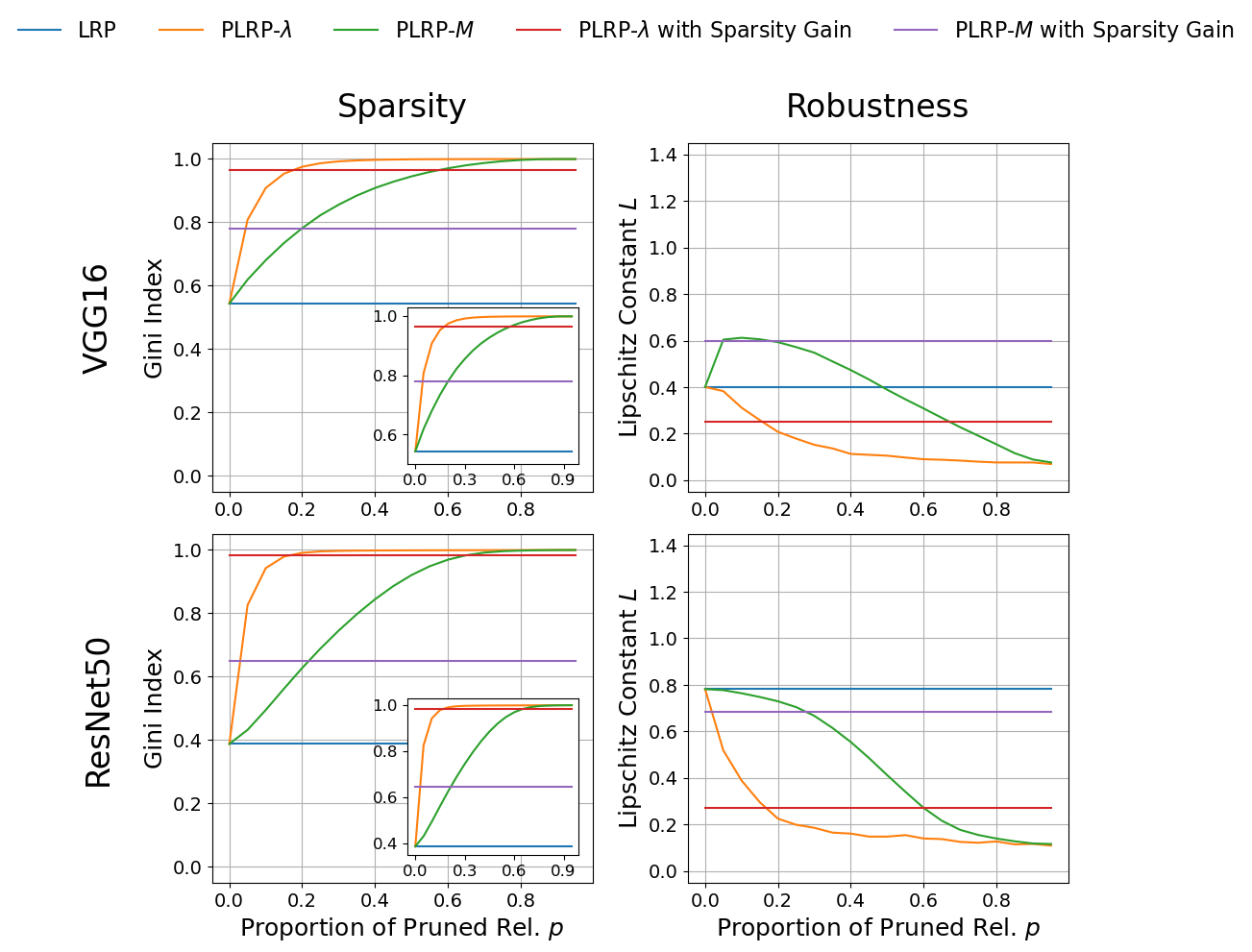}
	\caption{Results for ImageNet dataset for metrics sparsity and robustness for different proportions of pruned relevance mass $p$ ranging from 0 to 0.95 in increments of 0.05 for models VGG16 and ResNet50. For sparsity, the y-axis covers the whole output domain of $[0,1]$. The zoom plots focus on the actual covered output domain for better comparison of the methods.}
    \label{fig_imagenet_results}
\end{figure}

\begin{figure}[H]
	\centering
	\includegraphics[scale=0.25]{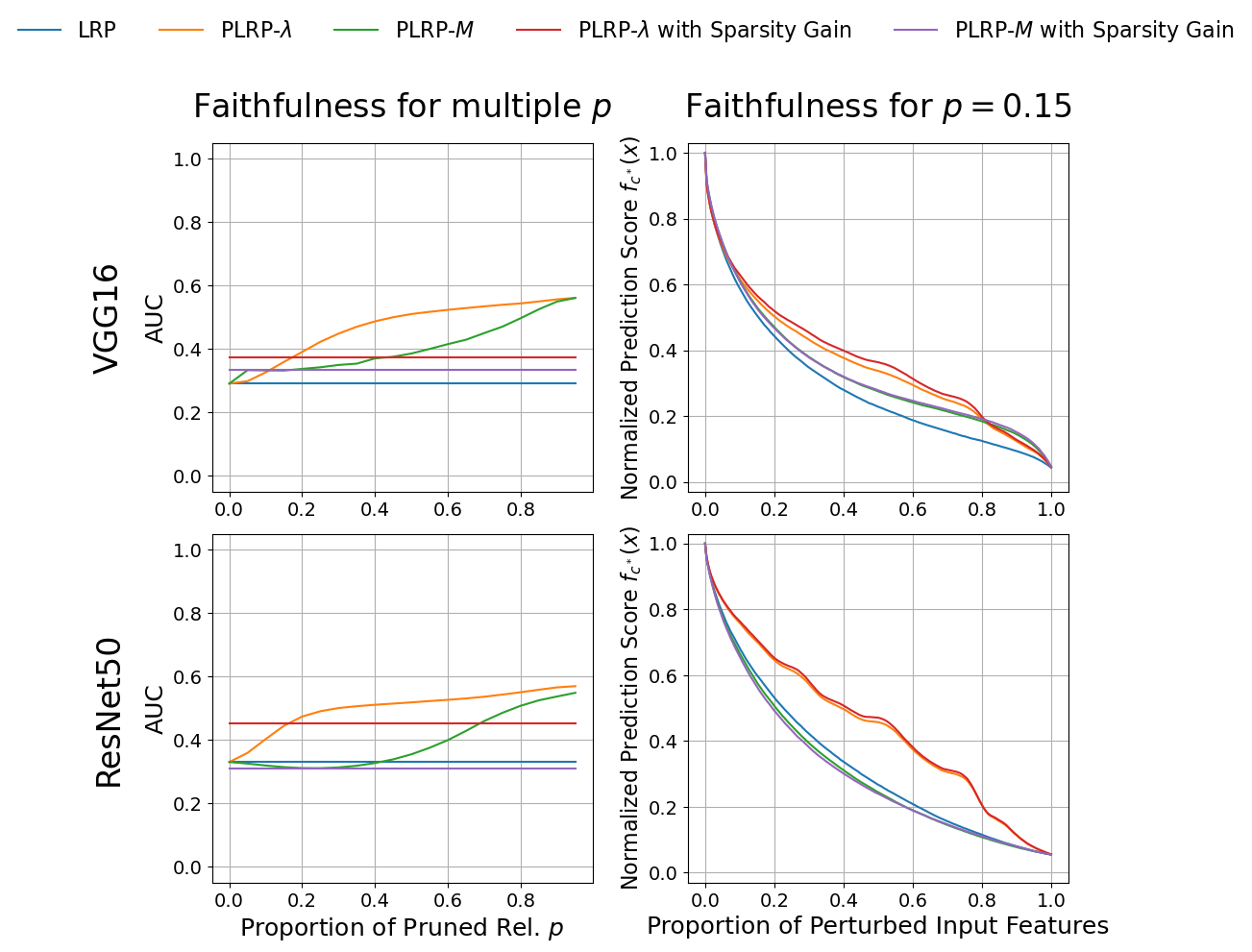}
	\caption{Results for ImageNet dataset for faithfulness for different proportions of pruned relevance mass $p$ ranging from 0 to 0.95 in increments of 0.05 (left) and for an exemplary parameterization $p=0.15$ (right) for models VGG16 and ResNet50. The AUC differs only slightly compared to the LRP baseline (left). For higher $p$, the difference becomes larger as more features are pruned. For the exemplary parameterization, the prediction score $f_{c^*}(\bx)$ drops similarly steeply as for the LRP baseline for the features with the highest relevance that are perturbed first (right). The difference in AUC is driven by the less important features that are perturbed later..}
    \label{fig_imagenet_results_faithfulness}
\end{figure}
\subsection*{B}
\begin{figure}[H]
	\centering
	\includegraphics[scale=0.25]{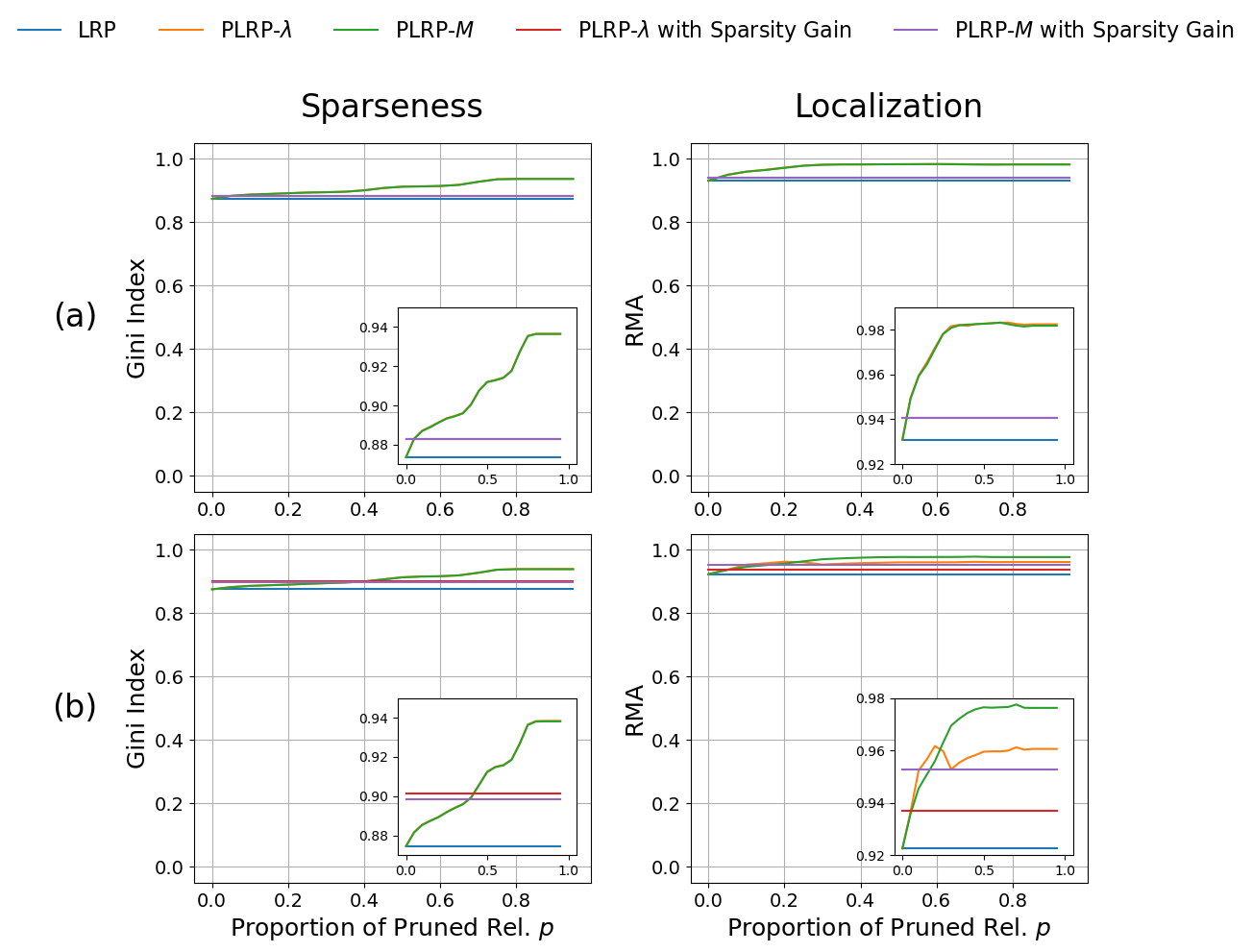}
	\caption{Results for genomics for metrics sparsity and localization for different proportions of pruned relevance mass $p$ for for a CNN with 32 filters (a) and 4 filters (b). The y-axis covers the whole output domain of $[0,1]$. The zoom plots focus on the actual covered output domain for better comparison of the methods.}
    \label{fig_genomics_results}
\end{figure}
%% ====================================================================================================================================================================================== %%
%% ====================================================================================================================================================================================== %%

\end{document}